\documentclass{l4dc2025}
\usepackage{siunitx}
\usepackage{setspace} % Add in the preamble if not already included
\usepackage{tikz}
\usepackage{bm}
\usepackage{lineno}
\usepackage[utf8]{inputenc}
\usepackage{diagbox}
\usepackage{array}% http://ctan.org/pkg/array (also necessary for width calculations (for vert. lines)))
\usepackage{wrapfig}
\usepackage{graphicx}
\usepackage{diagbox}
\usepackage{multirow}
\usepackage{caption}
\usepackage{soul}
\usepackage{enumitem}
\usepackage{xcolor}
\usepackage{colortbl}
\definecolor{lightgray}{gray}{0.92}
\usepackage{booktabs}
\usepackage{nameref}
\usepackage{booktabs}
\usepackage{enumitem} 
\usepackage{algorithm}
\usepackage{algpseudocode}
\usepackage{amssymb}
\usepackage{times}
\usepackage{array}
\usepackage{hyperref}
\usepackage{titlesec}
\titlespacing{\subsection}{0pt}{8pt}{*1}

\newcommand{\tripprox}{\setbox0\hbox{$\approx$}%
\mbox{\makebox[0pt][l]{\raisebox{0.48\ht0}{$\approx$}}$\approx$}}
\usepackage{multimedia}

\title[DKMGP for Vehicle Dynamics]{DKMGP: A Gaussian Process Approach to Multi-Task and Multi-Step Vehicle Dynamics Modeling in Autonomous Racing}

\author{%
 \Name{Jingyun Ning} \Email{jn2ne@virginia.edu}\\
 \addr  Department of Electrical \& Computer Engineering,
        University of Virginia
 \AND
 \Name{Madhur Behl} \Email{madhur.behl@virginia.edu}\\
 \addr Department of Computer Science, University of Virginia%
}

\begin{document}

\maketitle

\begin{abstract}
% Autonomous racing has garnered significant attention due to its potential to advance autonomous vehicle technologies. The performance of an autonomous racing car hinges on the seamless integration of perception, motion planning, and control models, particularly when operating at the limits of the vehicle’s capabilities. 
% Accurately modeling race car dynamics is crucial, as it not only captures the current state of the vehicle but also predicts future states, such as position, orientation, and velocity. 
Autonomous racing is gaining attention for its potential to advance autonomous vehicle technologies. Accurate race car dynamics modeling is essential for capturing and predicting future states like position, orientation, and velocity.
% Modeling individual subsystems, such as the engine, tires, suspension, and chassis, presents substantial challenges due to their complexity.
However, accurately modeling complex subsystems such as tires and suspension poses significant challenges.
% \hl{Traditional approaches, such as physics-based models and single-task Gaussian Processes, struggle with balancing prediction accuracy, computational efficiency, and ease of modeling, making them impractical for real-time control in autonomous racing.}
In this paper, we introduce the Deep Kernel-based Multi-task Gaussian Process (DKMGP), which leverages the structure of a variational multi-task and multi-step Gaussian process model enhanced with deep kernel learning for vehicle dynamics modeling. 
% The incorporation of deep kernel learning significantly enhances the model’s predictive performance, an essential required for model-based receding horizon control. 
% \hl{
Unlike existing single-step methods, DKMGP performs multi-step corrections with an adaptive correction horizon (ACH) algorithm that dynamically adjusts to varying driving conditions.
% }
To validate and evaluate the proposed DKMGP method, we compare the model performance with DKL-SKIP and a well-tuned single-track model, using high-speed dynamics data (exceeding \SI{230}{\kilo\meter\per\hour}) collected from a full-scale Indy race car during the Indy Autonomous Challenge held at the Las Vegas Motor Speedway at CES 2024.
The results demonstrate that DKMGP achieves upto 99\% prediction accuracy compared to one-step DKL-SKIP, while improving real-time computational efficiency by 1752x. 
%This balance of accuracy and efficiency makes DKMGP a practical choice for real-time applications in high-speed autonomous racing.
Our results show that DKMGP is a scalable and efficient solution for vehicle dynamics modeling making it suitable for high-speed autonomous racing control.

\end{abstract}

\section{Introduction}

As the technology of autonomous vehicles has developed during the past decade, there has been an increased growth of the research in autonomous racing (\cite{betz2022autonomous}, \cite{weiss2020deepracing}). 
Competitions in autonomous racing have been held not only in simulators (\cite{hartmann2021autonomous, babu2020f1tenth}), but also on hardware with racecars' ranging from 1:43 scale RC cars,
to 1/10 scale F1tenth Racing (\cite{o2019f1}) 
to full-size Indy racecars (\cite{carrau2016efficient, wischnewski2022indy}). 
To enable optimal performance, it is essential to learn an accurate vehicle dynamics model for model-based control algorithms. However, the nonlinear nature of vehicle subsystems makes constructing a high-fidelity dynamics model challenging, as first-principles approaches rely on complex mathematical equations to describe these dynamics (\cite{althoff2017commonroad}). Researchers often employ simplified vehicle models to overcome this complexity, such as single-track models. However, these models suffer from over-simplifications and do not fully capture vehicles' nonlinear behaviors.
In addition, first-principles modeling approaches are often costly and time-consuming to develop and fine-tune. For instance, the \cite{pacejka2005tire} tire model is commonly applied to capture tire-road interactions. Building this model requires parameters obtained from experimental data, such as grip, wear, and temperature changes, necessitating expert knowledge and specialized tire testing rigs to measure forces, torques, and displacements under various conditions.

\noindent \textbf{Related Works:} Consequently, researchers are increasingly turning to learning-based modeling approaches based on machine learning to bridge the discrepancy between simplified vehicle models and real dynamics (\cite{9357196, da2020modelling}).  \cite{pan2021data} proposed Deep Pacejka, a data-driven modeling approach based on deep neural networks (DNNs) for computing and predicting vehicle dynamics. \cite{chrosniak2024deep} has extended the Deep Pacejka by implementing the physics-informed neural network, which combines the physics coefficient estimation and dynamical equations to predict the vehicle dynamics of a full-sized racecar. Alternatively, hybrid models that combine physics-based insights with data-driven approaches offer a promising middle ground, leveraging the strengths of both methods. \cite{van2017online} first proposed Gaussian process (GP) models for learning vehicle dynamics, leveraging their interpretability and reliable uncertainty estimates.
The use of GP models to predict racecar dynamics by modeling the residuals between simplified models and real-world dynamics has been further investigated by researchers (\cite{jain2020bayesrace, kabzan2019learning}). The limitations of previous work of GP-based dynamics learning methods include unrealistic experiment setups, such as small-scaled racecars and the use of made-up racetracks (\cite{ning2024gaussian}).
% \hl{
While learning-based controllers using GPs are promising for capturing nonlinear dynamics, their high computational costs and latency, limit their integration into model-based controllers for high-speed applications like autonomous racing. 
Thus, there is a critical need for efficient learning-based models that can support real-time, closed-loop control strategies.
% }

\noindent \textbf{Contributions:} Building upon prior work, \cite{ning2023vehicle} proposed DKL-SKIP, which effectively predicts model errors in the dynamics of a full-size autonomous racecar. However, this method has two main limitations: it models each state separately, requiring three DNN models to predict all state residuals, and it operates only in a single-step prediction manner. 
In this paper, we extend the concept of DKL-SKIP and address its limitations by proposing the Deep Kernel-based Multi-task Gaussian Process (DKMGP) model. The contributions of this paper are as follows:

\begin{enumerate}
    \vspace{-8pt}
    \item We introduce the DKMGP, a multi-task variational Gaussian Process model that can predict all state residuals with a single correction model. 
    \vspace{-8pt}
    \item DKMGP can be trained for multi-step predictions, needed for closed-loop model-based control. Our second contribution is developing an adaptive correction horizon (ACH) algorithm to reduce inference times for multi-step predictions.
    % \item We develop a rule-based adaptive correction horizon algorithm aimed at reducing the inference times for the DKMGP to improve its suitability for real-time applications further.
    \vspace{-8pt}
    \item We validate DKMGP using high-speed racing data from a full-scale autonomous racecar from the Indy Autonomous Challenge, reaching speeds over \SI{230}{\kilo\meter\per\hour}. 
%and compare its performance to a physics-based nonlinear single-track model and DKL-SKIP.
\end{enumerate}
While the DKMGP model presented in this work focuses on enhancing vehicle dynamics prediction and real-time performance; both of these are critical precursors for model-based closed-loop control in autonomous racing. 
To the best of our knowledge, this is the first work to utilize a learning-enabled multi-task Gaussian Process for vehicle dynamics modeling for autonomous racing. 
%While this paper focuses on dynamics modeling, the efficiency and accuracy of DKMGP lay the groundwork for integrating learning-based models with closed-loop controllers.

\section{Background and Problem Formulation}
%\input{modeling}
%Before formulating the problem that integrates the Extended Kinematic (E-kin) model with the proposed DKMGP model to enhance dynamics modeling accuracy, 
We first present a brief overview of the single-track and the E-kin models used in this paper. 
\setlength{\belowcaptionskip}{0pt}
\setlength{\textfloatsep}{8pt}
\begin{table*}[]
    \scriptsize
    \centering
    \caption{E-kin model and single-track model state equations; highlighted states differ between the models, including the Pacejka tire model equations}
    \resizebox{\textwidth}{!}{
    \begin{tabular}{l l l l}
        \hline
        \textbf{State Variable} & \textbf{Notation} & \textbf{State Equation (E-Kin)} & \textbf{State Equation (Single-track)} \\
         Horizontal Position (\si{\meter}) & $x$ & $\dot x = v_{x}\cos\psi - v_{y}\sin\psi$ & $\dot x = v_{x}\cos\psi - v_{y}\sin\psi$ \\
         Vertical Position (\si{\meter}) & $y$ & $\dot y = v_{x}\sin\psi_t + v_{y}\cos\psi$ & $\dot y = v_{x}\sin\psi + v_{y}\cos\psi$ \\
\rowcolor{lightgray}         
Longitudinal Velocity (\si{\meter\per\second}) & $v_x$ & {$\dot v_{x}$} $= \frac{T_w}{l_r+l_f}a_x$ & {$\dot v_{x}$} $= a_x$ \\[3pt]
\rowcolor{lightgray}           
Lateral Velocity (\si{\meter\per\second}) & $v_y$ & {$\dot v_{y}$} $= \frac{1}{T_w(l_r+l_f)}(a_x\sin\psi+ v_x\omega)$ & {$\dot v_{y}$} $= \frac{1}{m}(F_{ry}+F_{fy}\cos\delta-F_{by})- v_x\omega$ \\
       
Inertial Heading (\si{\radian}) & $\psi$ &  $\dot \psi = \omega$ & $\dot \psi = \omega$ \\[3pt]
         Steering Angle (\si{\radian})& $\delta$ & $\dot \delta = \Delta\delta$ & $\dot \delta = \Delta\delta$ \\
\rowcolor{lightgray}            
 Yaw Rate (\si{\radian\per\second}) & $\omega$ & {$\dot \omega$} $= \frac{h_{cog}}{T_w(l_r+l_f)}(a_x\cos\psi+ v_x\omega)$ & {$\dot \omega$} $= \frac{1}{I_z}(l_{f}F_{fy}\cos\delta-l_{r}F_{ry})$ \\[3pt]
         
        \hline
        \textbf{Tire Forces (Pacejka)} & & & \\
        Front Lateral Tire Force (\si{\newton}) & $F_{yf}$ & \multicolumn{2}{l}{$F_{yf} = Svy_{f} + D_f \sin( C_f \arctan ( B_f \alpha_f - E_f( B_f \alpha_f  -\arctan(B_f \alpha_f))))$} \\[3pt]
        Rear Lateral Tire Force (\si{\newton}) & $F_{yr}$ & \multicolumn{2}{l}{$F_{yr} = Svy_{r} + D_r \sin( C_r \arctan( B_r \alpha_r - E_r( B_r \alpha_r - \arctan(B_r \alpha_r))))$} \\[3pt]
        Front Slip Angle (\si{\radian}) & $\alpha_f$ & \multicolumn{2}{l}{$\alpha_f = \alpha_{f0} + Shy_{f}, \quad \alpha_{f0} = \delta - \arctan \left(\frac{v_y + l_f \omega}{v_x}\right)$} \\[3pt]
        Rear Slip Angle (\si{\radian}) & $\alpha_r$ & \multicolumn{2}{l}{$\alpha_r = \alpha_{r0} + Shy_{r}, \quad \alpha_{r0} = -\arctan \left(\frac{v_y - l_r \omega}{v_x}\right)$} \\
        \hline
    \end{tabular}
    }
    \label{tab:dynamics_equations}
\end{table*}
\subsection{Single-track Model}
\begin{wrapfigure}{r}{0.4\textwidth}
    \centering
    \includegraphics[width=0.4\textwidth]{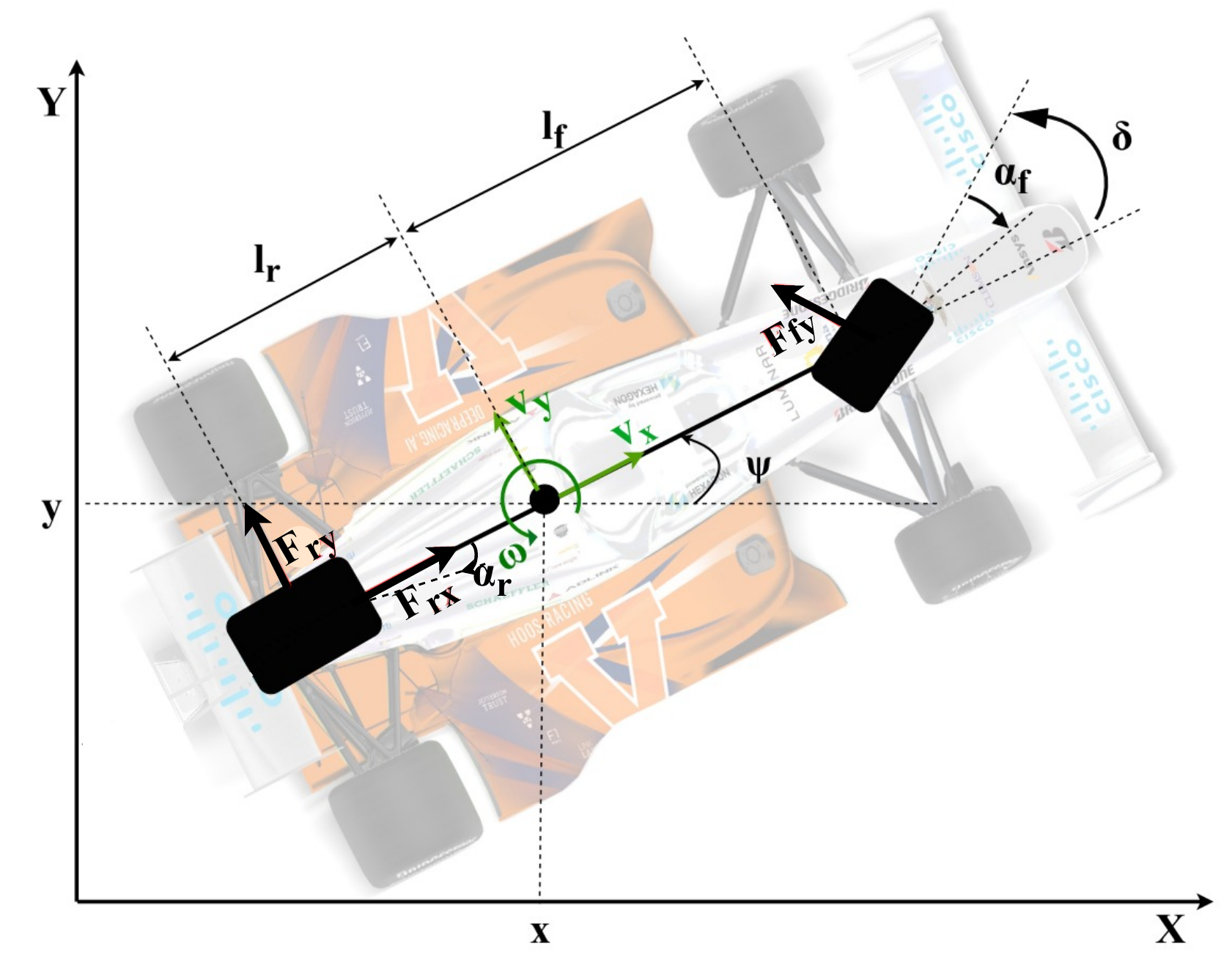}
    \caption{The single-track vehicle model of the racecar, with all three base states $(v_x. v_y, \omega)$ highlighted.}
    \vspace{-2pt}
    \label{fig:single-track}
\end{wrapfigure}

The single-track model, as shown in Fig.\ref{fig:single-track}, simplifies the vehicle dynamics by combining the two wheels of both the front and rear axles into a single wheel per axle. This model is commonly used for motion planning and real-time vehicle control (\cite{raji2022motion, liniger2015optimization}.) The differential equations of state descriptions of the single-track model are shown in the Table~\ref{tab:dynamics_equations}, where the vehicle state is composed of $s=\{x, y, v_x, v_y, \psi,$ $\delta, \omega\}$, and the inputs are $u=\{a_x, \Delta\delta\}$, where $\Delta\delta$ represents the steering velocity, $a_x$ is the longitudinal vehicle acceleration, respectively.
%Specifically, the single-track model used in this paper assumes that 
% The longitudinal dynamics of the racecar are characterized directly by the vehicle's longitudinal acceleration, $a_x$. 
Note that in this work, we simplify the longitudinal dynamics by directly using the vehicle's longitudinal acceleration, whereas in real-world applications, a well-calibrated drivetrain model is necessary to precisely compute the force transferred from the engine to the driving wheels.
The lateral dynamics of the racecar are characterized by states including lateral velocity ($v_y$), yaw rate ($\psi$), steering angle ($\delta$), and yaw rate ($\omega$). 
The associated differential equations incorporate forces such as the lateral tire forces of the front ($F_{fy}$) and rear ($F_{ry}$) axles, and the force due to bank angle ($F_{b_y}$), where $F_{b_y} = mg \sin(\theta)$ approximates the influence of the bank angle $\theta$. Tire forces, $F_{fy}$ and $F_{ry}$, are modeled using the Pacejka magic formula for the tire model, detailed in the bottom of Table~\ref{tab:dynamics_equations}. 

\subsection{Extended Kinematic Model}
\setlength{\intextsep}{3pt} % Reduce vertical space above and below
\setlength{\columnsep}{8pt} % Reduce horizontal space to the text
Table~\ref{tab:dynamics_equations} demonstrates that all vehicle states can be derived from the base states \( \{v_x, v_y, \omega\} \). While the single-track model simplifies vehicle dynamics modeling, it still requires extensive expert knowledge and calibration. For instance, deriving \( \dot{v}_y \) involves lateral dynamics influenced by aerodynamics and tire forces, while \( \dot{v}_x \) requires understanding of the powertrain system.
To reduce model development costs, we propose an extended kinematic single-track (E-kin) model, shown in column 3 of Table~\ref{tab:dynamics_equations}. Unlike the nonlinear single-track model, which relies on detailed subsystem modeling, the E-kin model approximates the derivatives of the base states using measurable variables such as \( a_x \), \( \psi \), and \( \delta \). This approach avoids explicit modeling of complex subsystems like tire dynamics and drivetrain dynamics, greatly simplifying and reducing modeling costs.
However, the E-kin model’s simplifications introduce discrepancies between its predictions and real-world vehicle dynamics. Thus, a method is needed to learn and compensate for these mismatches while retaining the model’s simplified calibration benefits.
\subsection{Problem Formulation}
\label{sec:ms}
% To measure the discrepancies between the E-kin model and the actual racecar dynamics, 
Given a dataset \( D =\{d_1, d_2, \dots, d_N\}\) consisting of $N$ vehicle state measurement samples \( S = \{s_1, s_2, \dots, s_N\} \) and inputs \( U = \{u_1, u_2, \dots, u_N\} \), the objective is to predict future states $\hat{s}_{t+1}, \ldots, \hat{s}_{t+n}$ using a learned model that corrects for the inaccuracies in the baseline E-kin model.

\textbf{Problem 1: Model Error Correction} 
The dynamics predicted by a simplified model, $\hat{s}_{t+1}$, are generated by propagating the E-kin model with the inputs \( u_t =  \{a_{x,t}, \Delta\delta_t\} \) and initial states \( s_t = \{x_t, y_t, v_{x,t}, v_{y,t}, \psi_t, \delta_t, \omega_t\} \). 
These predictions, that we will denote with $f_{\text{E-kin}}(s_t, u_t)$, deviate from the true system behavior $s_{t+1}$, due to unmodeled complexities.
Define the model residual at time $t+1$ as $r_{t+1} = s_{t+1} - f_{Ekin}(s_{t}, u_{t})$, where $r_{t+1} = [0, 0, \epsilon_{v_x,t+1}, \epsilon_{v_y,t+1}, 0, 0, \epsilon_{\omega,t+1}]$ captures the errors $\epsilon$ in predicting the base dynamic states $\{v_x, v_y, \omega\}$ at time \( t+1 \).
Our goal is to learn a model $f_{\text{corr}}$ that corrects the baseline predictions using these residuals:
\setlength{\abovedisplayskip}{5pt}
\setlength{\belowdisplayskip}{5pt}
\begin{equation}
    f_{\text{corr}}(s_{t+1}, u_{t+1}) = f_{\text{E-kin}}(s_{t}, u_{t}) + \mathbf{e}_{t+1},
\end{equation}
where $\mathbf{e}_t = \{\epsilon_{v_x,t}, \epsilon_{v_y,t}, \epsilon_{\omega,t}\}$ represents the estimated model error, $r_{t}$, which is estimated using a Gaussian Process (GP) model:  $\mathbf{e}_t = \mathcal{GP}(d_{t-1}) + \nu_t, \quad \nu_t \sim \mathcal{N}(0, \sigma^2)$, with the previous state measurement $d_{t-1}$ as the input. 
Existing GP-based models, such as DKL-SKIP (\cite{ning2023scalable}), predict each correction state independently, resulting in high computational costs due to the space and time complexity of GP models. 
The challenge in this paper is to use a single correction model to improve the suitability of the method for closed-loop control. We leverage correlations between the base state variables $\{v_x, v_y, \omega\}$ using a Multi-Task Gaussian Process (MTGP) model, that predicts all model errors simultaneously.

\textbf{Problem 2: Multi-Step Prediction}
% Given an input sequence $\{s_t, u_t\}_{t=1}^{n}$, the objective in multi-step prediction is, at time \( t \), to predict multiple future states over a horizon $n$:
The objective in multi-step prediction is, at time $t$, to predict future states over a given prediction horizon $m$ i.e.
\setlength{\abovedisplayskip}{5pt}
\setlength{\belowdisplayskip}{5pt}
\begin{equation}
\begin{array}{ccc}
    \hat{s}_{t+1} = f_{Ekin}(s_t, u_t); & 
    \dots; & 
    \hat{s}_{t+m} = f_{Ekin}(\hat{s}_{t+m-1}, u_{t+m-1}) \\
\end{array}
\label{eq:ekin_prop}
\end{equation}

% If you keep propagating the E-Kin model auto-regressively the prediction at the end of the horizon $\hat{s}_{t+n}$ will deviate significantly from the true state $s_{t+n}$. 
However, auto-regressive propagation of the E-Kin model leads to significant deviations, between the prediction $\hat{s}_{t+m}$ and the true state $s_{t+m}$, at the end of the prediction horizon.
Additionally,  one can keep correcting the model prediction at each step of the horizon but that is computationally prohibitive. 
Therefore, we need to learn an MTGP model that can correct the deviations at fixed intervals that we term as the correction horizon, $n$, for the MTGP model, with $n<m$.
% Alternatively, one can keep correcting the model prediction at each step of the horizon but that is computationally prohibitive. 
The MTGP model needs to directly correct the $n^{th}$ prediction, following $(n-1)$ E-Kin model propagations:
\begin{equation}
    \hat{s}_{t+n} = f_{\text{E-kin}}(\hat{s}_{t+n-1}, u_{t+n-1}) + \mathbf{e}_{t+n},
\end{equation}
$\mathbf{e}_{t+n}$ is estimated using the DKMGP framework presented next. 
Choosing good values for the correction horizon $n$, itself is also a problem of interest for our work.

\section{DKMGP Methodology}
\begin{figure}[h]
    \centering
    \begin{minipage}{0.49\linewidth}
        \centering
        \includegraphics[width=\linewidth]{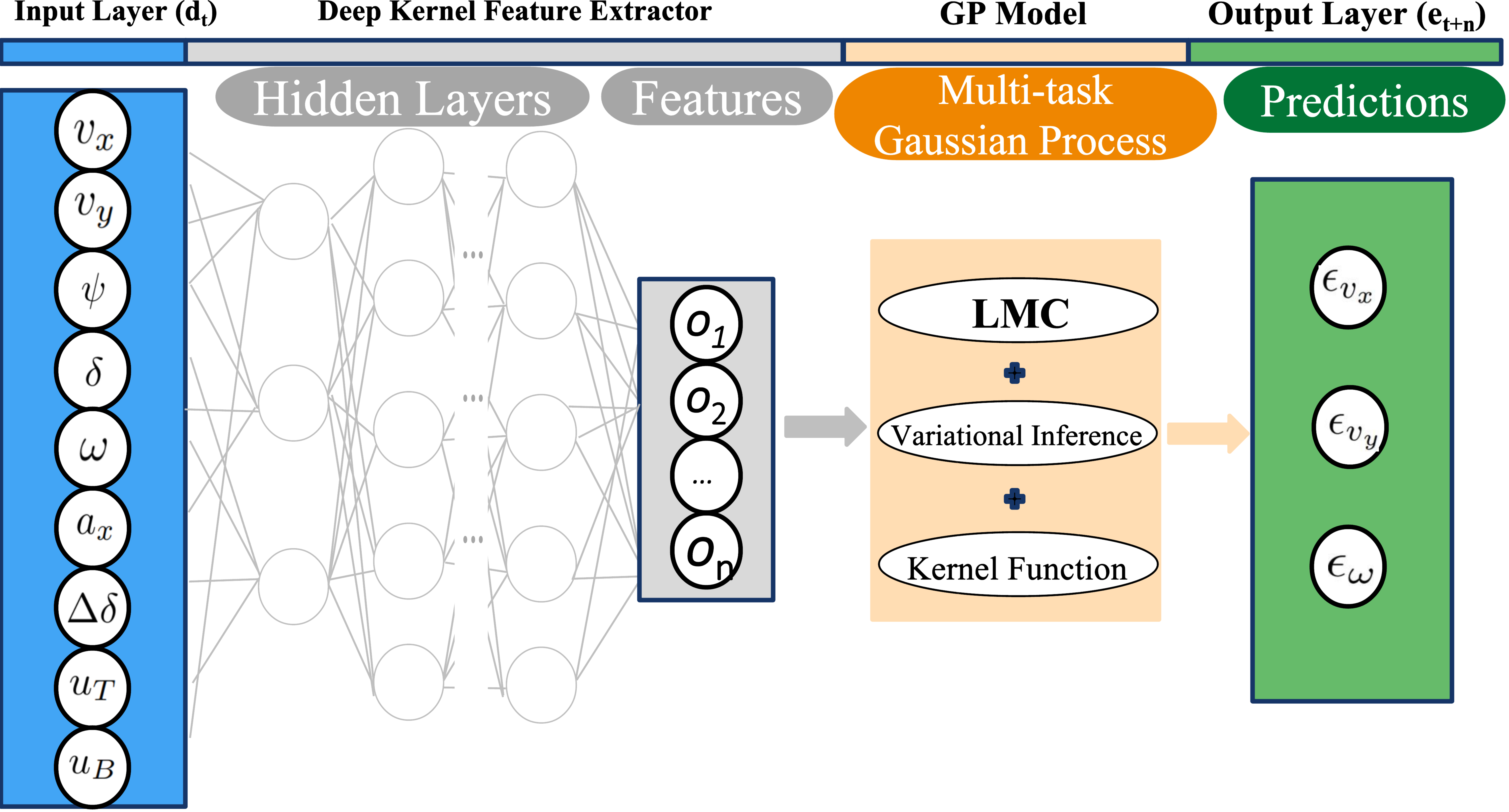}
        \caption{The architecture of the DKMGP model.}
        \label{fig:multigp-block1}
    \end{minipage}
    \hfill % Adds horizontal space between the figures
    \begin{minipage}{0.49\linewidth}
        \centering
        \includegraphics[width=\linewidth]{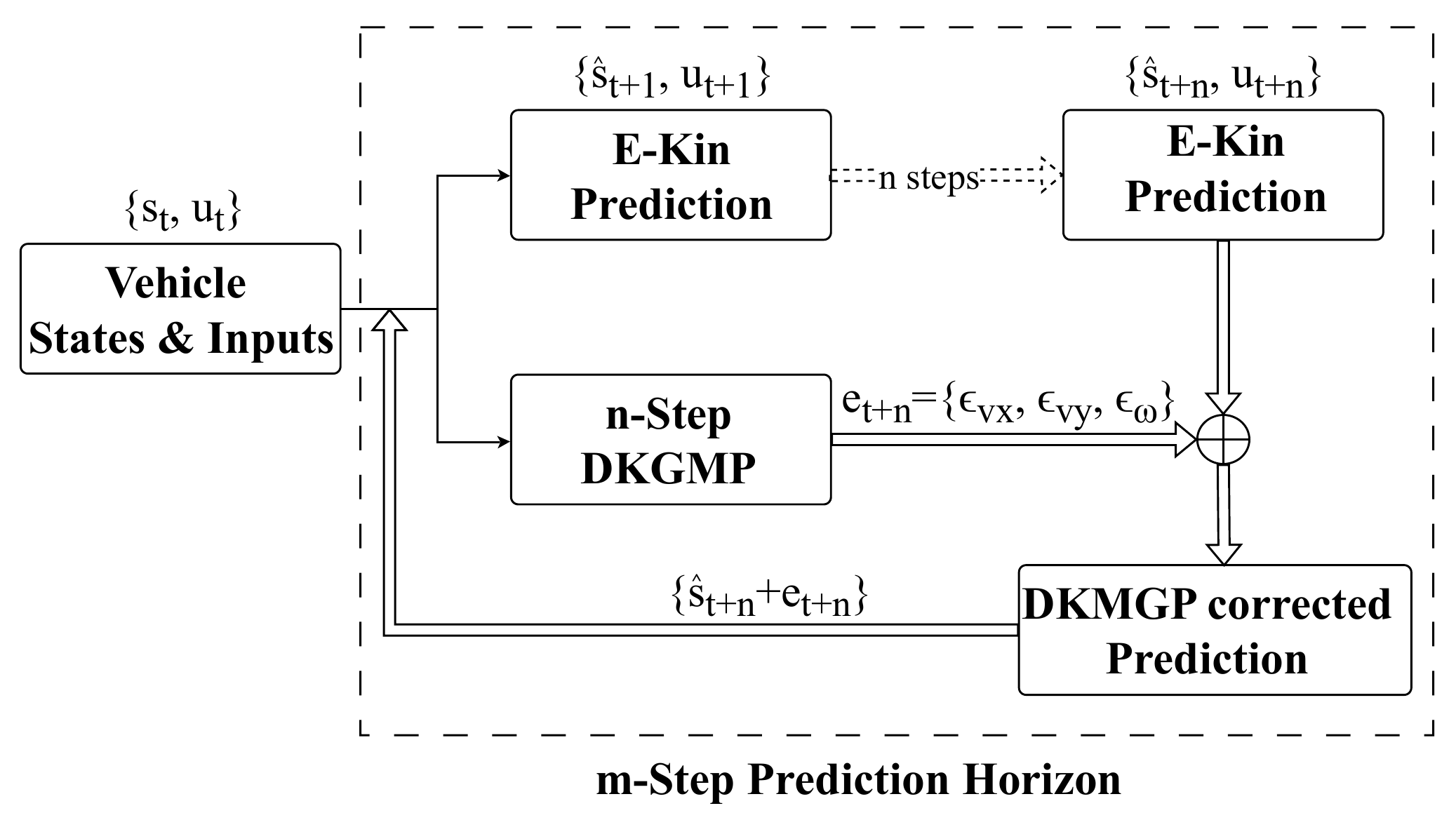}
        \caption{Workflow of the multi-step approach for the DKMGP predictions.}
        \label{fig:multi-pred}
    \end{minipage}
\end{figure}

This section details the methodology of the proposed DKMGP model, depicted in Figure~\ref{fig:multigp-block1}.
The model integrates a deep neural network (DNN) with a multi-task variational Gaussian Process (MTGP). The DNN serves as a feature extractor, adept at capturing non-linear relationships and extracting relevant information from the input data. 
The MTGP employs a Linear Model of Coregionalization 
(LMC) within a variational framework to predict multiple targets simultaneously. 
% Together, these components aim to efficiently model the discrepancies between the E-kin predictions and the actual racecar dynamics.
\setlength{\intextsep}{5pt} % Reduce vertical space above and below
\setlength{\columnsep}{8pt} % Reduce horizontal space to the text

\subsection{Deep Kernel Learning}
Gaussian process regression, as a non-parametric method, does not learn new representations of the data. Instead, it utilizes a predetermined kernel function to measure the similarity between data points (\cite{rasmussen2003gaussian}). Specifically, rather than transforming or adapting the input data into a new feature space, as seen in the neural network models, GP defines a fixed feature space determined by the selected kernel function and its hyperparameters. 
Therefore, the limited expressiveness and adaptability of GP methods make it challenging for them to capture complex relationships between variables in highly nonlinear and dynamic environments, such as autonomous racing as shown in previous work of \cite{ning2024gaussian}.
% Besides, GP does not involve transforming or adapting the input data into a new feature space through training, as seen in models like neural networks. Rather, the kernel defines a feature space implicitly, which is fixed based on the chosen kernel and its parameters, not learned from the data. 
To address this, we use the deep kernel learning (DKL) method aiming to transform a standard kernel \(k(d_i, d_j|\theta)\) to a deep kernel through a neural network, $g(\cdot)$, as denoted in Equation~\ref{eq:deepK}.
\vspace{-4pt}
\begin{equation}
    k(d_i, d_j;\theta) \rightarrow k(g(d_i,w),g(d_j,w)|\theta,w) \tripprox k(\tilde{d}_i, \tilde{d}_j; \gamma)
    \label{eq:deepK}
\end{equation}
Here, \( g(d, w) \) denotes a deep neural network with weights \( w \), while \( \theta \) represents the hyperparameters of GP kernel, such as the length-scale \( l \) in an RBF kernel. To learn data representations, the collective hyperparameters for the deep kernel, denoted as \( \gamma = \{\theta, w\} \), can be jointly learned and optimized.
% by maximizing the log marginal likelihood, as detailed in Equation~\ref{eq:mll}.
% \begin{equation}
% \log p(\mathbf{y} | \gamma, X) \propto - \left[ \mathbf{y}^T (K_{\gamma} + \sigma^2 I)^{-1} \mathbf{y} + \log | K_{\gamma} + \sigma^2 I | \right]
% \label{eq:mll}
% \end{equation}
Furthermore, since DKL serves as a feature extractor, its transformation enables DKL to capture the most relevant features from the input data while simultaneously reducing its dimensionality. Specifically, it maps each input sample \( d \in \mathbb{R}^9 \) to a lower-dimensional feature representation \( \tilde{d} \in \mathbb{R}^4 \), thereby reducing the computational burden of GP models, as expressed in Equation~\ref{eq:deepK}. 

% \begin{equation}
%     k(d_i, d_j; \theta) \rightarrow k(\tilde{d}_i, \tilde{d}_j; \gamma)
%     \label{eq:deepK2}
% \end{equation}

\subsection{Multi-Task Gaussian Process}
Different from single-task Gaussian Process, in the context of the MTGP model, we consider a dataset \( D \) comprising \( N \) distinct inputs, denoted as \( \{d_1, \dots, d_N\} \), and associated with these inputs are outcomes corresponding to Multiple tasks. There are three tasks in this paper, denoted as $\{\epsilon_{v_x}, \epsilon_{v_y}, \epsilon_\omega\}$, which are derived from the model residuals of the base states. These tasks are represented as \( \{\epsilon_{v_{x,1}}, \dots, \epsilon_{v_{x,N}}; \epsilon_{v_{y,1}}, \dots, \epsilon_{v_{y,N}}; \epsilon_{\omega_1}, \dots, \epsilon_{\omega_N}\} \), forming an \( T \times N \) matrix, where $T=3$ and each element corresponds to the response of a task to the \( i^{th} \) input \( d_i \).

\noindent \textbf{Linear model of coregionalization}
Given that the model residuals of the vehicle base states \{$v_x$, $v_y$, $\omega$\} are correlated, where changes in one state affect the others due to the physical interactions between the vehicle’s dynamics, we apply the linear model of coregionalization (LMC) method to capture these correlations across the tasks. 
The goal of LMC is to leverage the correlation across different tasks to improve prediction accuracy and learning efficiency, thus providing a robust framework for MTGP. 
Instead of assuming that each task is independent, LMC models the tasks as being correlated through one or more latent functions. 
For instance, LMC assumes that the $T$ task-specific functions $f_1, f_2, \dots, f_T$ are generated from a smaller number of latent function $h_1, \dots, h_Q$, which induce correlations between the tasks (\cite{alvarez2012kernels}). 
Therefore, for the dataset with $T$ tasks and $Q$ latent functions, the MTGP for the $\tau$-th task at input $d_i$ is expressed as: $f_\tau(d_i) = \sum_{q=1}^Q a_{\tau q} h_q(d_i) + \nu_\tau$.
where \( h_q \) are GP latent functions with zero mean and covariance \( h_q(d_i) \sim \mathcal{GP}(0, k_q(d_i, d_j)) \). Task similarities are established through the mixing coefficients \( a_{\tau q} \), which determine how the latent functions contribute to each task's output. 
\( \nu_\tau \) represents independent noise for the \( \tau \)-th task. 
The covariance between any two tasks \( \tau \) and \( \tau' \) at inputs \( d_i \) and \( d_j \) is given by: $Cov(f_\tau(d_i), f_{\tau'}(d_j)) = \sum_{j=1}^q a_{\tau q} a_{\tau 'q} k_q(d_i,d_j)$.
% \begin{equation}
%     Cov(f_m(d_i), f_{m'}(d_j)) = \sum_{j=1}^q a_{mq} a_{m'q} k_q(d_i,d_j)
%     \label{eq:lcm_cov}
% \end{equation}
Therefore, under the LMC framework and the assumption of $Q$ independent latent GP functions, the vector-valued function $F(d_i)=[f_1(d_i); ...; f_\tau(d_i)]$ follows a GP prior defined as:$F(d_i) \sim \mathcal{GP}(0, \sum_{q=1}^Q A_q k_q(d_i,d_j))$.
% \begin{equation}
%     F(d_i) \sim \mathcal{GP}(0, \sum_{q=1}^Q A_q k_q(d_i,d_j))
%     \label{eq:lcm_gp}
% \end{equation}
Where $A_q \in R^{T \times T}$ is known as the coregionalization matrix, and elements of each $A_q$ are the coefficients $a_{\tau q} a_{\tau'q}$.

\noindent \textbf{Variational inference}
The computational cost grows to $\mathcal{O}((T \cdot N)^3)$ for MTGP, compared to $\mathcal{O}(N^3)$ for single-task GP due to the involvement of multiple tasks, $T$. This cubic scaling significantly increases the computational burden, making direct computation of the true posterior distribution prohibitively expensive in MTGP. To overcome this, we use variational inference (VI) to approximate the true posterior distribution with a simpler, tractable distribution $q(\mathbf{f})$. 
The objective of VI optimization is to minimize the Kullback-Leibler (KL) divergence between the approximations, $\text{KL}(p(\mathbf{\epsilon_\tau}|\mathbf{f})||q(\mathbf{f}))$, which measures how closely $q(\mathbf{f})$ approximates $p(\mathbf{\epsilon_\tau}|\mathbf{f})$ (\cite{wainwright2008graphical}). This optimization is achieved by maximizing the evidence lower bound (ELBO), derived from the marginal likelihood and given by Equation~\ref{eq:elbo}.
\vspace{-4pt}
\begin{equation}
    \text{ELBO} = \mathbb{E}_{q(\mathbf{f})}[\log p(\mathbf{\epsilon_\tau}|\mathbf{f})] - \text{KL}(q(\mathbf{f}) || p(\mathbf{f}))
\label{eq:elbo}
\end{equation}
Here, $\mathbb{E}_{q(\mathbf{f})}[\log p(\mathbf{\epsilon_\tau}|\mathbf{f})]$ represents the expected log-likelihood under the variational distribution, and $\text{KL}(q(\mathbf{f}) || p(\mathbf{f}))$ is the KL divergence between the variational distribution and the prior.

\subsection{DKL Integration}
While MTGP is powerful for modeling correlations across multiple tasks and capturing shared structures, it relies on predefined kernels that may be struggling to handle highly complex, non-linear relationships present in real-world data. This limitation becomes especially evident when dealing with large datasets and high-dimensional inputs. Previous work has demonstrated the effectiveness of the integration of GP with DKL, which addresses the limitations by leveraging DNN to learn highly expressive representations that can benefit GP framework (\cite{ning2023scalable, wilson2016deep}). The DKL transformation from input space to feature space benefits MTGP from both the flexibility of deep learning to extract meaningful features and the robustness of GP-based uncertainty quantification and multi-task correlation modeling. Besides, this transformation improves the scalability of MTGP to high-dimensional data, as shown in Equation~\ref{eq:deepK}, the original data $d_i$ is transformed to a lower-dimensional feature representation, $\tilde{d}_i$. 
% \begin{figure}[] % First figure
%     \centering
%     \includegraphics[width=0.4\linewidth]{images/mtgp_block1.pdf}
%     \caption{The architecture of the DKMGP model.}
%     \label{fig:multigp-block1} % Correct label after caption
% \end{figure}%
% \begin{figure}[] % Second figure
%     \centering
%     \includegraphics[width=0.4\linewidth]{images/multi-step.pdf}
%     \caption{Workflow of the multi-step approach for the DKMGP prediction of vehicle dynamics.}
%     \label{fig:multi-pred}
% \end{figure}

DKL with MTGP allows for the simultaneous learning of both feature extraction and GP parameters in an end-to-end framework. This is achieved by jointly optimizing the ELBO along with the deep kernel function, which is parameterized by the neural network. The joint optimization problem involves maximizing the ELBO with respect to both the variational parameters, $\lambda$, and the DKL parameters, $\gamma$, as shown in Equation~\ref{eq:eblodkl}. Gradients are computed for the neural network parameters and are used to update both the deep neural network and the variational parameters simultaneously.
\begin{equation}
\text{ELBO}(\lambda, \gamma) = \mathbb{E}_{q(\mathbf{f};\lambda)}[\log p(\mathbf{\epsilon_\tau}|\mathbf{f}, \gamma)] - \text{KL}(q(\mathbf{f};\lambda) || p(\mathbf{f}|\gamma))
\label{eq:eblodkl}
\end{equation}
By combining the strengths of DNN and MTGP, the DKMGP model provides a highly flexible and scalable approach for complex multi-task learning challenges. This makes it well-suited for modeling vehicle dynamics, especially in highly nonlinear racing environments where the interactions between multiple dynamic factors are crucial for performance.

\section{Multi-step DKMGP}
To address Problem 2 in section~\ref{sec:ms}, we propose a multi-step prediction approach, focusing on the adaptive correction horizon algorithm.

\noindent \textbf{Training Phase:}
% The training phase for DKMGP is performed offline, where the model learns residuals between the baseline E-kin predictions and actual dynamics data.
The goal of the DKMGP training phase is to learn the residuals between the baseline E-kin predictions and actual dynamics data offline, enhancing model efficiency during the inference phase.
In order to train the DKMGP model for multi-step prediction with a given horizon $n$, we initialize the E-kin model with recorded states, \( s_t \), and propagate it \( n-1 \) steps using the input sequence \( \{u_t, u_{t+1}, \dots, u_{t+n-1}\} \). 
We then measure the residual errors $e_{t+n}$ between the E-kin predictions, $\hat{s}_{t+n}$, and ground truth, $s_{t+n}$. 
The DKMGP model is trained to predict this error, using initial measurements, $s_{t}$, as input features and errors \( e_{t+n} \) as target outputs. 
The model’s kernel choice, batch size, and learning rate were optimized via a grid search, see section~\ref{sec:hyp_tun}.

\noindent \textbf{Inference Phase:}
Figure~\ref{fig:multi-pred} shows the workflow during inference time. At time \( t \), initial vehicle states \( s_t \) and inputs \( u_t \) are measured and fed into both the DKMGP and E-kin models. 
The E-kin model is then propagated over $n-1$ steps without correction, and the DKMGP will predict the errors between E-kin predictions and actual states at the \( n \)-th step. 
After that, the DKMGP corrects the E-kin predictions.  
Notably, the length of correction horizon $n$ can adapt dynamically throughout the prediction horizon $m$.
% \hl{Here, the correction horizon ($n$) is shorter than the full prediction horizon ($m$).} 
This multi-step prediction and correction cycle iterates over longer horizons, with each corrected prediction, \( \hat{s}_{t+n} + e_{t+n} \), feeding back as the initial state for the next cycle. 
This process repeats until the prediction horizon \( m \) is reached, after which the actual measured states and inputs, \( \{s_{t+m}, u_{t+m}\} \), are reinitialized. 
This iterative approach ensures that predictions remain accurate and computationally efficient over time.
\begin{algorithm}[]
\scriptsize
\setstretch{1} % Reduce the line spacing to 80%
\SetKwInOut{Input}{Input}
\SetKwInOut{Output}{Output}
\caption{Adaptive Correction Horizon (ACH) for DKMGP}
\label{alg:ACH_algorithm}
\Input{velocity $v_x$[\SI{}{\meter\per\second}], acceleration $a_x$[\SI{}{\meter\per\second^2}], steering wheel angle $\delta_w$[\SI{}{\degree}]}
\Output{Correction horizon $n$ (steps)}
% \vspace{-5pt} % Reduce the space after input/output
\textbf{Step 1: Classification of input variables}
\begin{itemize}[noitemsep, topsep=0pt] % Remove itemize spacing
    \item Classify $v_x$, $a_x$, and $\delta_w$ based on thresholds:
    \begin{itemize}[noitemsep, topsep=0pt] % Remove inner itemize spacing
        \item \textbf{Cruising}: $v_x \in [0, 40)$, $|a_x| \in [0, 0.5)$, $|\delta_w| \in [0, 4.5)$
        \item \textbf{Controlled}: $v_x \in [40, 50)$, $|a_x| \in [0.5, 1.0)$, $|\delta_w| \in [4.5, 7.5)$
        \item \textbf{Pushing}: $v_x \in [50, 60)$, $|a_x| \in [1.0, 3.0)$, $|\delta_w| \in [7.5, 11.5)$
        \item \textbf{Aggressive}: $v_x \geq 60$, $|a_x| \geq 3.0$, $|\delta_w| \geq 11.5$
    \end{itemize}
\end{itemize}
% \vspace{-5pt} % Reduce space between steps
\textbf{Step 2: Determine overall driving condition}
\begin{itemize}[noitemsep, topsep=0pt]
    \item Select the most aggressive classification across $v_x$, $a_x$, and $\delta_w$
\end{itemize}
\textbf{Step 3: Map to correction horizon}
\begin{itemize}[noitemsep, topsep=0pt]
    \item Assign $n$ based on condition:
    \begin{itemize}[noitemsep, topsep=0pt]
        \item Cruising $\rightarrow$ 15 steps, Controlled $\rightarrow$ 10 steps, Pushing $\rightarrow$ 5 steps, Aggressive $\rightarrow$ 3 steps
    \end{itemize}
\end{itemize}
\textbf{Step 4: Output correction horizon $n$}
\end{algorithm}

When determining the length of the correction horizon, $n$, several factors must be considered. 
Shorter horizons improve accuracy by reducing deviations but introduce more computational overhead. 
In contrast, longer horizons are computationally efficient but can lead to less precise predictions. 
To balance this trade-off, we present a rule-based adaptive correction horizon (ACH) algorithm that dynamically adjusts the horizon length based on real-time driving conditions, such as speed, acceleration, and steering angle.
The ACH algorithm, detailed in Algorithm~\ref{alg:ACH_algorithm}, categorizes the racecar's driving conditions into four levels: \textit{Cruising}, \textit{Controlled}, \textit{Pushing}, and \textit{Aggressive}. 
Based on these classifications, the algorithm sets the correction horizon to one of the predefined lengths: 3, 5, 10, or 15 steps. 
The real-time driving condition is represented using three measurements: longitudinal velocity \( v_x \), acceleration \( a_x \), and steering wheel angle \( \delta_w \).
The correction horizon length is determined by selecting the most aggressive classification among these factors.
Therefore, ACH enables the DKMGP model to dynamically adjust its correction horizon and adapt to changing driving conditions.

\section{Experiments and Results}
\setlength{\intextsep}{-1pt} % Reduce vertical space above and below
\setlength{\columnsep}{8pt} % Reduce horizontal space to the text
\begin{wrapfigure}{r}{0.5\textwidth}
    \centering
    \includegraphics[width=0.5\columnwidth]{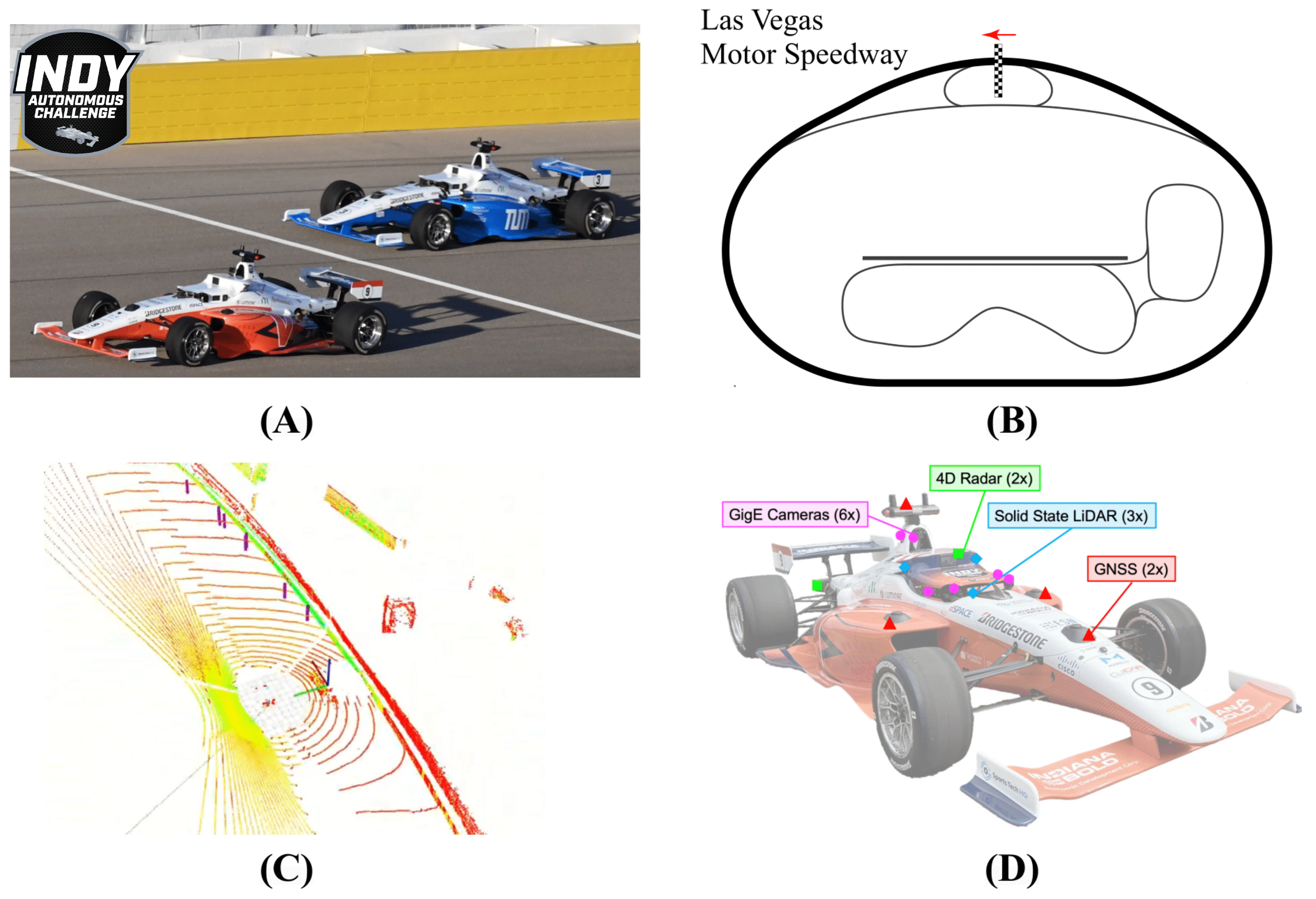}
    \caption{(A) Data from overtaking competition, LVMS CES 2024. (B) LVMS track layout. (C) LiDAR point cloud data. (D) AV-21 sensor setup.}
    \label{fig:ov}
\end{wrapfigure}
This section outlines the experimental setup and validation results for the DKMGP-integrated E-kin model, focusing on its prediction accuracy and real-time performance.
% against the DKL-SKIP method as well as a nonlinear single-track model.

\subsection{Real-world Data Collection}
We evaluate the performance of the DKMGP model using real-world autonomous racing data collected from AV-21, a full-scale, fully autonomous racecar that competes in the Indy Autonomous Challenge (IAC). 
AV-21 is equipped with LiDAR, GNSS, cameras, radar, and advanced computing hardware essential for autonomous operation, as shown in Fig. \ref{fig:ov}(D).
The data was collected during an autonomous passing competition held at the Las Vegas Motor Speedway (LVMS) during the 2024 IAC CES event, as shown in \ref{fig:ov}. In this data, our racecar achieved a top speed of \SI{230}{\kilo\meter\per\hour}. 
State estimations obtained using an EKF algorithm are logged at \SI{100}{\hertz} and recorded as ROS2 bag files. 
These bag files (similar to RACECAR~\cite{kulkarni2023racecar} dataset) are processed to construct the dataset, $D_{real}$. 

\setlength{\intextsep}{0pt} % Reduce vertical space above and below
\setlength{\columnsep}{5pt} % Reduce horizontal space to the text
\begin{wraptable}{r}{0.55\textwidth}
\centering
\scriptsize
\setlength{\tabcolsep}{2pt}  % Reduce column spacing
\renewcommand{\arraystretch}{1.5}  % Adjust row spacing
\begin{tabular}{|l|l|}
\hline
\multicolumn{1}{|c|}{\textbf{Hyperparameters}} & \multicolumn{1}{c|}{Range ; \textbf{Optimal Values}} \\ \hline
Batch Size & 80 to 280 ; \textbf{144} \\ \hline
Epochs & 400 to 1300 ; \textbf{1140}\\ \hline
Kernel Functions & SM, RQ, RBF, Polynomial, Periodic, Matérn; \textbf{RBF} \\ \hline
Hidden Layer 1 Size & 128 to 512 neurons; \textbf{256} \\ \hline
Hidden Layer 2 Size & 32 to 64 neurons; \textbf{64} \\ \hline
Learning Rate & 0.005 to 0.1; \textbf{0.0064} \\ \hline
Number of Features & 3 to 5; \textbf{5} \\ \hline
\end{tabular}
\caption{Hyperparameter Ranges and Optimal Values}
\label{tab:ranges}
\end{wraptable}
This data is split into a training set and a testing set. The training set contains data from approximately 17 laps (\SI{1410}{\sec}), while the testing set, which remains unseen by the model, consists of data from a different set of another 17 laps (\SI{1417}{\sec}). 

\subsection{DKMGP Hyperparameters Tuning}
\label{sec:hyp_tun}
We performed extensive hyperparameter tuning and ablation study for the DKMGP Model, optimizing parameters 
such as batch size, epoch count, kernel type, DNN hidden layer sizes, learning rate, and the number of features from the DKL feature extractor. 
Using the Optuna API (\cite{akiba2019optuna}), we conducted 400 trials on an NVIDIA GeForce RTX 3080 GPU, with 32 GB RAM and an Intel Core i7 processor. Weights \& Biases (\cite{wandb}) tracked the hyperparameter configurations and metrics for analysis. 
The ranges and optimal values for each hyperparameter are presented in detail in Table~\ref{tab:ranges}.

\subsection{DKMGP Model Validation}
% We evaluate DKMGP model's performance in prediction accuracy and computational efficiency.
\setlength{\intextsep}{-1pt} % Reduce vertical space above and below
\setlength{\columnsep}{5pt} % Reduce horizontal space to the text
\setlength{\abovecaptionskip}{0pt} % Example adjustment
\setlength{\belowcaptionskip}{2pt}
\begin{wraptable}{r}{0.55\textwidth}
\centering
\scriptsize
\setlength{\tabcolsep}{2pt}  % Reduce column spacing
\renewcommand{\arraystretch}{1.5}  % Adjust row spacing
\begin{tabular}{|c|c|c|c|c|c|}
\hline
\textbf{Fixed Horizon} & \textbf{n3} & \textbf{n5} & \textbf{n10} & \textbf{n15} & \textbf{DKL-SKIP} \\ \hline
\textbf{Inf. Rate} & \SI{22.45}{\hertz} & \SI{35.38}{\hertz} & \SI{61.35}{\hertz} & \SI{75.33}{\hertz} & \SI{0.043}{\hertz} \\ \hline
\end{tabular}
\caption{Inference rates for DKMGPs and DKL-Skip over a 43-step prediction horizon}
\label{tab:inference_rates}
\end{wraptable}
\noindent \textbf{DKMGP real-time performance}
We use the same hardware setup described in Section \ref{sec:hyp_tun} to evaluate the DKMGP model efficiency. Additionally, we implement the DKL-SKIP approach under identical conditions to provide a comparison with the proposed DKMGP method. As shown in Table~\ref{tab:inference_rates}, the inference rate of the DKMGP models, without code optimization, achieves inference rates between \SI{22.45}{\hertz} (n3) and \SI{75.33}{\hertz} (n15), which are approximately 522 to 1752 times faster than the DKL-SKIP rate of \SI{0.043}{\hertz}, over a total 43-step prediction horizon. 
% Specifically, the DKL-SKIP inference rate is as low as \SI{0.043}{\hertz}, which is unsuitable for real-time control applications. 
The inefficiency of DKL-SKIP arises from its use of three separate GP models to correct for each error term, $\{\epsilon_{v_x}, \epsilon_{v_y}, \epsilon_\omega\}$, individually, combined with its single-step prediction approach. In contrast, the DKMGP model simultaneously predicts all error terms over multiple steps, leading to a more efficient inference process. 
% This approach results in a substantial performance improvement. 
% \setlength{\intextsep}{0pt} % Reduce vertical space above and below
% \setlength{\columnsep}{5pt}
% \setlength{\abovecaptionskip}{5pt} % Example adjustment
% \setlength{\belowcaptionskip}{-2pt}
% \begin{wrapfigure}{r}{0.5\textwidth}
% \centering
%     \includegraphics[width=0.5\textwidth]{images/ach_pred.pdf}
%     \caption{Comparison of DKMGP models across different fixed correction horizons.}
%     \label{fig:ach_fix}
% \end{wrapfigure}
Consequently, these results demonstrate that the DKMGP model offers a significant enhancement over DKL-SKIP in real-time performance for vehicle dynamics modeling.

\begin{minipage}[c]{0.48\textwidth}
    \centering
    \scriptsize
    \renewcommand{\arraystretch}{1.2}  % Adjust row spacing
    \begin{tabular}{|c|c|c|c|c|}
        \hline
        \multirow{2}{*}{Step} & \multirow{2}{*}{Metric} & \multicolumn{3}{c|}{Error Type} \\
        \cline{3-5}
        & & $\mu_{v_x}$ & $\mu_{v_y}$ & $\mu_\omega$ \\
        \hline
        \multirow{3}{*}{DKL-SKIP} 
        & MAE & 0.1311 & 0.0533 & 0.0218 \\
        & RMSE & 0.2341 & 0.0835 & 0.0315 \\
        \hline
        \multirow{3}{*}{DKMGP} 
        & MAE & 0.1351 & 0.0943 & 0.0424 \\
        & RMSE & 0.2300 & 0.1272 & 0.0547 \\
        \hline
        \multirow{2}{*}{E-kin ST} 
        & MAE & 0.2521 & 1.0084 & 0.5868 \\
        & RMSE & 0.3059 & 1.6511 & 0.9708 \\
        \hline
    \end{tabular}
    \captionsetup{width=0.9\textwidth}
    \captionof{table}{Comparison of prediction errors across different models.}
    \label{tab:evaluation}
\end{minipage}
% \hfill
\begin{minipage}[c]{0.48\textwidth}
    \centering
    \setlength{\abovecaptionskip}{-1pt}
    \includegraphics[width=\textwidth]{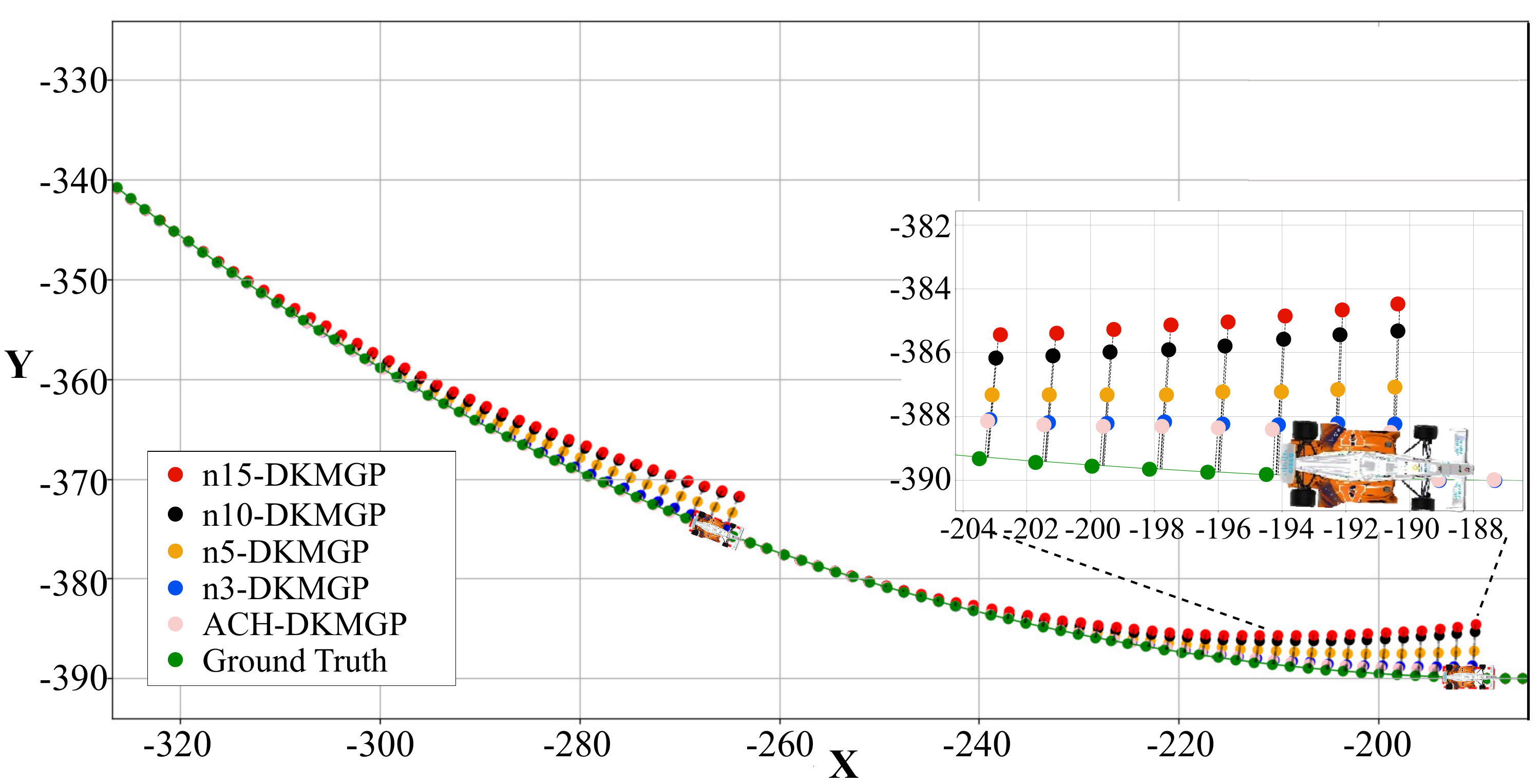}
    \captionsetup{width=0.9\textwidth}
    \captionof{figure}{Comparison of DKMGP models across different fixed correction horizons.}
    \label{fig:ach_fix}
\end{minipage}
\vspace{1pt}

\noindent \textbf{DKMGP prediction accuracy} 
We first compare the ACH-based DKMGP model to DKMGP models with fixed-length correction horizons. Figure~\ref{fig:ach_fix} shows prediction results across a 43-step horizon (\SI{1.7}{\sec}), specifically assessing cross-track errors (CTE) by comparing model predictions to ground truth. The ACH-based DKMGP consistently achieves the smallest CTE, aligning closely with ground truth measurements. Additionally, the DKMGP model’s predictions remain within half a car length, minimizing the risk of track boundary collisions.
Then, we compare the prediction accuracy of DKMGP to the DKL-SKIP method-based E-kin model, all benchmarked against an uncorrected E-kin model using mean absolute error (MAE) and root mean squared error (RMSE) as metrics. 
The inference rate of DKL-SKIP (\SI{0.043}{Hz}) was too slow to evaluate its performance on the entire dataset, and therefore, its validation was restricted to just a single lap from the data with average values of base states being \(v_x = \SI{54.5}{\meter\per\second}\), \(v_y = \SI{0.31}{\meter\per\second}\), and \(\omega = \SI{0.19}{\radian\per\second}\). 
As shown in Table~\ref{tab:evaluation}, the DKL-SKIP achieves slightly higher accuracy than the DKMGP model, with relative errors lower by  0.07\%, 13.23\%, and 10.79\%  of the averages for \(v_x\), \(v_y\), and \(\omega\), respectively. 
However, this minor accuracy gain comes with a significant trade-off in computational efficiency as due to DKL-SKIP's low inference rate.
Notably, DKMGP demonstrates substantial improvements over the uncorrected E-Kin model, especially with relative errors in \(v_y\), and \(\omega\) that are 294.19\%, and 286.32\% lower, respectively. 

\subsection{Comparison of DKMGP and Nonlinear Singe-track}
\begin{wrapfigure}{r}{0.5\textwidth}
    \centering
    \includegraphics[width=0.51\columnwidth]{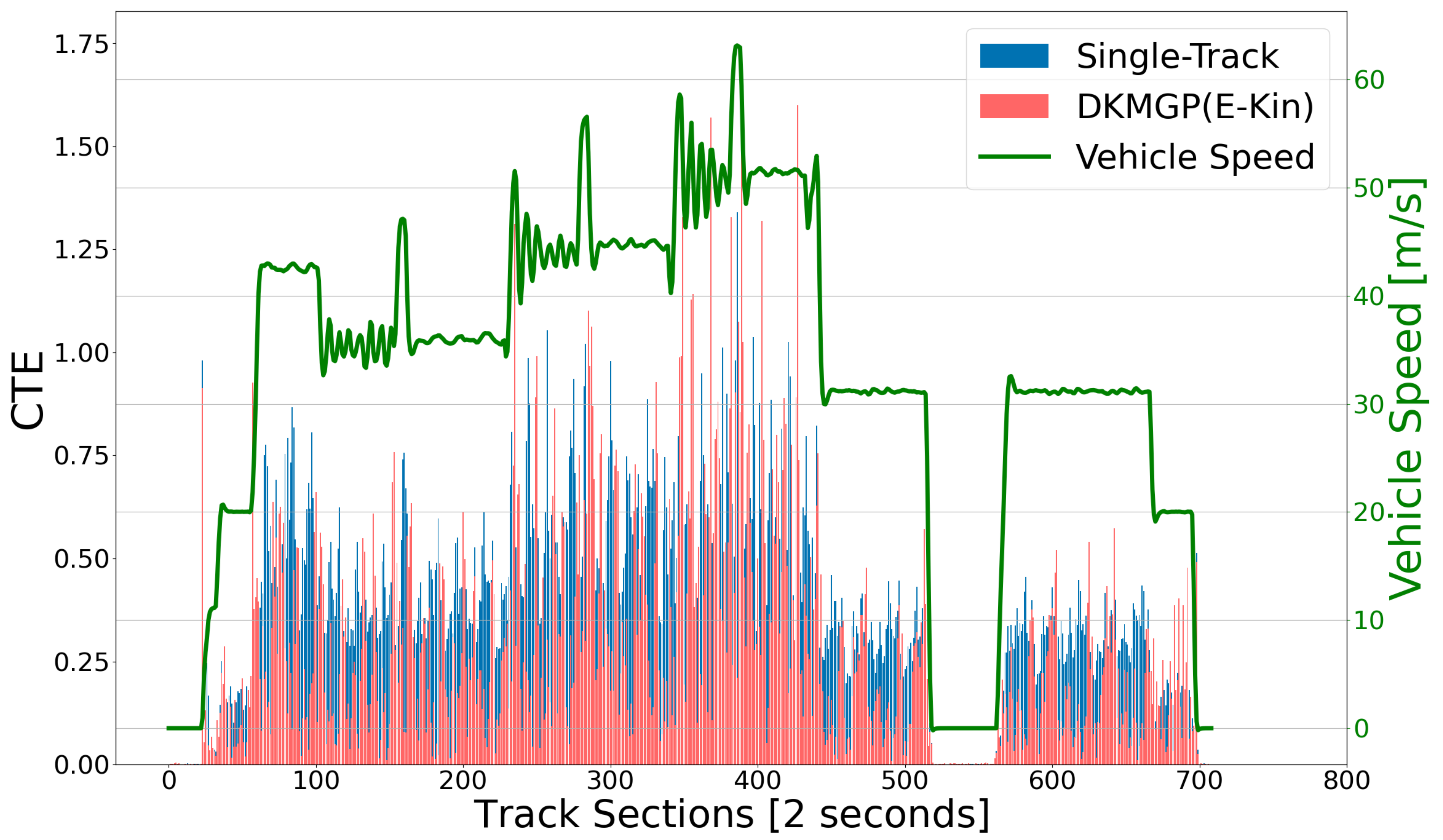}
    \caption{Comparison of sectional cross-track errors between the single-track model and the DKMGP (E-kin) model.}
    \label{fig:sectional}
\end{wrapfigure}

Finally, we compare the DKMGP-based E-kin model with a nonlinear single-track model, which has been validated as effective for modeling vehicle dynamics at high speeds exceeding \SI{230}{\kilo\meter\per\hour} in autonomous racing. 
% It is important to note that developing this model required years of data gathering and extensive fine-tuning of parameters across various subsystems. 
% Tire and drivetrain parameters were tuned using manufacturer-provided data, and dyno testing (expensive).
% We present this comparison with the intention of evaluating the accuracy of DKMGP to a well-tuned physics-based model, noting that in practice, such a model would be very challenging to obtain.
Developing this model required years of data collection and extensive parameter tuning across subsystems, including tuning tire and drivetrain parameters using manufacturer-provided data, and dyno testing (expensive).
We compare DKMGP with this well-tuned physics-based model to evaluate its accuracy, acknowledging that such a model would be challenging to obtain in practice.
We use sectional average cross-track error (sCTE) as the metric, with each section covering a 2-second interval. This segmentation allows for detailed analysis of prediction accuracy across track sections while minimizing the impact of short-term outliers. 
Figure~\ref{fig:sectional} shows the sCTE comparison between the DKMGP-based E-kin model (red) and the single-track model (blue).
\setlength{\intextsep}{2pt} % Reduce vertical space above and below
\setlength{\columnsep}{5pt} % Reduce horizontal space to the text
The E-kin model consistently achieves lower sCTE below \SI{40}{\meter\per\second}, showing better prediction performance over the single-track model under these conditions. 
This advantage indicates that the E-kin model’s adaptive correction capabilities allow it to closely follow ground truth measurements. 
At higher speeds above \SI{45}{\meter\per\second}, the single-track model excels due to its ability to account for slip angle effects in vehicle dynamics. 
Notably, the E-kin model is still able to maintain accuracy within one car width from ground truth, which is within the margin of safety typically used in autonomous racing. 
These findings demonstrate that with significantly lower effort in model building and tuning, the E-kin model, together with DKMGP, can outperform the single-track model at low to moderate speeds while achieving comparable performance at higher speeds. 
% These results highlight that, with significantly less effort required for building and calibration, the E-kin model can outperform the single-track model at low to medium speeds while delivering comparable performance at high speeds.

\section{Conclusions}
In this paper, we developed a DKMGP-based E-kin model to capture the dynamics of a full-size autonomous racecar. 
The DKMGP model overcomes the limitations of existing methods by predicting all state residuals using a single model.  
Additionally, we present an adaptive correction horizon algorithm that enables DKMGP for multi-step prediction. 
Evaluation with real-world data from the Indy Autonomous Challenge showed that DKMGP achieves comparable accuracy to DKL-SKIP, while providing orders of magnitude higher computational efficiency than DKL-SKIP, with inference rates $\sim 500$ to 1700 times faster, making it suitable for model-based predictive control.
Above all, the DKMGP combined with E-kin model, requires significantly less development effort, compared to a nonlinear single-track model. Future work will focus on implementing and evaluating DKMGP within a closed-loop controller, such as model predictive control, and incorporating the uncertainty provided by GP models to improve controller performance and safety.

% Future work will explore closed-loop evaluation using advanced control algorithms, such as model predictive control, and will further optimize the method for robust performance across diverse racing conditions.

% In this paper, we propose a DKMGP-based extended kinematic single-track (E-kin) model to capture the dynamics of a full-size autonomous racecar. The DKMGP model addresses limitations in the DKL-SKIP method by predicting all dynamic residuals within a multi-step framework. Evaluation of high-speed racing data from the Indy Autonomous Challenge demonstrated that the DKMGP E-kin model achieves high prediction accuracy, is suitable for real-time use, and requires significantly less development effort than a nonlinear single-track model. The E-kin model outperformed the single-track model in non-aggressive conditions and remained competitive at high speeds.

% Future work will explore closed-loop evaluation using advanced control algorithms, such as model predictive control, and will further optimize the method for robust performance across diverse racing conditions.

\section*{Supplementary Demo Video}
A demo reel showcasing the Deep Kernel-based Multi-task Gaussian Process (DKMGP) in action, including real-time vehicle dynamics prediction and its application in high-speed autonomous racing, is available at \url{https://youtu.be/q6qFsDyoGqQ}

% \clearpage
% \bibliographystyle{plain} 
\bibliography{references}

\end{document}